\documentclass[10pt]{article} 
\usepackage[accepted]{tmlr}

\usepackage{amsmath,amsfonts,bm}

\def\eqref#1{equation~\ref{#1}}

\def\1{\bm{1}}

\DeclareMathAlphabet{\mathsfit}{\encodingdefault}{\sfdefault}{m}{sl}
\SetMathAlphabet{\mathsfit}{bold}{\encodingdefault}{\sfdefault}{bx}{n}

\usepackage{hyperref}
\usepackage{url}
\usepackage{amsmath,amssymb}
\usepackage{booktabs}
\usepackage{graphicx}
\usepackage[table]{xcolor}
\usepackage{multirow}
\usepackage{array}
\usepackage{longtable}
\usepackage{placeins}
\setcitestyle{numbers,square,comma,sort&compress}

\definecolor{omegaDecode}{HTML}{2F6DB5}
\definecolor{omegaDeploy}{HTML}{C0603A}
\definecolor{omegaAccent}{HTML}{E4B23C}
\definecolor{omegaMute}{HTML}{9A9A9A}

\title{Reversion, Retention, and Semantic Intervention: Measuring Functional Content in CLIP Fine-Tuning Deltas}

\author{\name Chih-Ting Liao\thanks{Correspondence to: Chih-Ting Liao (\texttt{mill.liao@unsw.edu.au}).} \email mill.liao@unsw.edu.au \\
      \addr University of New South Wales
      \AND
      \name Kezhen Bao \email kezhen.bao@student.unsw.edu.au \\
      \addr University of New South Wales
      \AND
      \name Xin Cao \email xin.cao@unsw.edu.au \\
      \addr University of New South Wales}

\def\month{10}
\def\year{2026}
\def\openreview{\url{https://openreview.net/forum?id=AQ7a5vGI0v}}

\begin{document}

\maketitle

\begin{abstract}
Fine-tuning adapts pretrained models to new tasks, but identifying which weight changes support the resulting behavior remains difficult. Weight-space interpretability assigns semantic descriptions to these changes, called fine-tuning deltas. Such descriptions leave open how much task performance depends on the identified updates, particularly when other updates can support the same predictions. We address this gap by pairing reversion, which restores selected parameters to their pretrained values, with retention, which keeps only selected fine-tuning updates. We evaluate Contrastive Language--Image Pretraining (CLIP) models on synthetic financial charts, Flowers102, EuroSAT, and FGVCAircraft with ViT-L/14, and on Flowers102 with ViT-B/16, using a shared recipe that tunes the final six visual blocks. Our contributions are threefold. First, attention value--output (VO) updates alone recover 43--87\% of the accuracy improvement, although reverting them leaves chart accuracy unchanged. Cross-entropy loss preserves this contrast on Charts and EuroSAT, while correct-class score margins reveal additional contributions. Second, VO weights recover more loss improvement than all ten equally sized, layer-matched random subsets in each setting. VO and multilayer perceptron (MLP) retentions jointly correct 25--81\% of pretrained errors, demonstrating overlapping recovered behavior, which we call operational redundancy. Third, removing semantically selected directions reduces VO-retained recovery more than matched-random removal in all five settings. Together, these results distinguish semantic alignment, removal cost, and retained recovery as complementary evidence about the functional role of fine-tuning updates within the evaluated CLIP recipe.
\end{abstract}

\section{Introduction}
\label{sec:intro}
Fine-tuning adapts a pretrained model by changing its parameters. Understanding which changes support the acquired behavior matters for interpreting adaptation and for choosing weights to edit or remove. The difference between the fine-tuned and pretrained parameters is the \emph{fine-tuning delta}. Weight-space interpretability analyzes this difference to identify structures associated with the new task, giving researchers a way to study adaptation directly in the model's parameters.

Semantic interpretation provides one such view. The semantic weight-reading method SITH \citep{gentile2026sith} assigns text descriptions to directions in attention value--output (VO) operators, which combine the value and output projections of an attention head. Applied to fine-tuning deltas, it reveals task-aligned structure even when the dominant directions of the absolute weights change little. Task singular vectors also represent adaptation for model merging \citep{gargiulo2025tsv}. These descriptions raise a functional question: how much of the acquired performance is supported by the updates that the readout identifies?

Removing selected updates answers only part of this question. The remaining updates may support the same predictions, so a small performance drop does not reveal how much improvement the removed updates could recover when retained. This distinction matters in transformers: attention and multilayer perceptron (MLP) updates are trained jointly and act in series through the residual stream, the shared representation passed between blocks. Accuracy alone can also conceal changes in prediction confidence. Interpreting an update's functional role therefore requires specifying both the surrounding network and the performance measure.

We pair two interventions to separate these questions. \emph{Reversion} restores selected parameters to their pretrained values while leaving the other updates in place; its performance cost measures necessity under that edit. \emph{Retention} keeps only selected fine-tuning updates and restores the rest; its recovery measures the fraction of the full fine-tuning improvement regained through the shared pretrained network. We compare both interventions in accuracy, cross-entropy loss, and correct-class margin, the score of the correct class minus the highest competing class score. Random-subset controls test whether recovery depends on the chosen parameters, and shared corrections of pretrained errors measure behavioral overlap.

\begin{figure}[t]
\centering
\includegraphics[width=\textwidth]{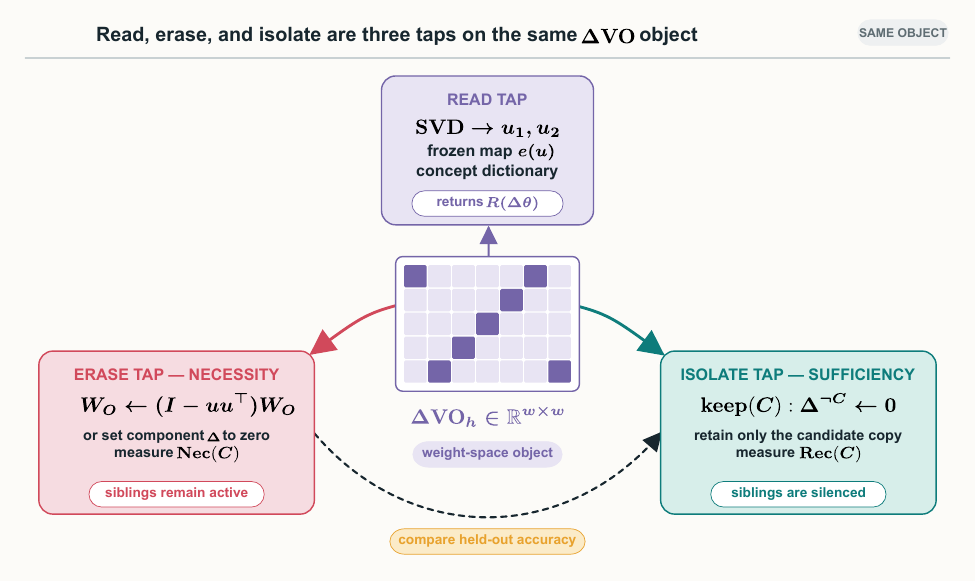}
\caption{\textbf{Semantic reading, reversion, and retention answer different questions.} The semantic readout acts on per-head effective VO deltas. Reversion measures the effect of removing selected updates from the tuned network; component retention measures recovery when the complementary updates are reverted. Both networks retain the pretrained computation. The schematic uses ``copy'' as shorthand for retained recovery. Direction removal is a targeted intervention on the readout, whereas the retention evidence is at component level.}
\label{fig:taps}
\end{figure}

We study Contrastive Language--Image Pretraining (CLIP) \citep{radford2021clip}, which maps images and text to a shared embedding space, in five classification setups: synthetic financial charts, Flowers102 \citep{nilsback2008flowers}, EuroSAT \citep{helber2019eurosat}, and FGVCAircraft \citep{maji2013aircraft} on ViT-L/14, plus Flowers102 on ViT-B/16. These span chart-pattern recognition, flower classification, satellite-image land-use classification, and fine-grained aircraft recognition. ViT-L/14 and ViT-B/16 are the large and base vision-transformer variants, with 14- and 16-pixel patches, respectively. All runs fine-tune the final six visual blocks while keeping class-text embeddings and the visual projection fixed. The class-text embeddings serve as classification targets, which we call text anchors. Charts allow us to control the concepts used in training; the standard datasets test whether the behavioral contrast extends beyond that construction.

On Charts, accuracy rises from 0.136 to 0.985 after tuning. Reverting the VO updates leaves accuracy unchanged, whereas retaining VO alone recovers 0.808 of the improvement. MLP-only retention recovers 0.989. Across the five settings, VO and MLP retention recover 0.428--0.869 and 0.682--0.989 respectively. Necessity costs vary across settings: Aircraft VO costs 6.1 accuracy points, and Aircraft and ViT-B/16 MLP reverts cost 12.7 and 12.4 points. We therefore use \emph{operational redundancy} for overlapping recovered behavior, and distinguish low-necessity cases from partial retention with appreciable necessity costs.

\begin{figure}[t]
\centering
\includegraphics[width=\textwidth]{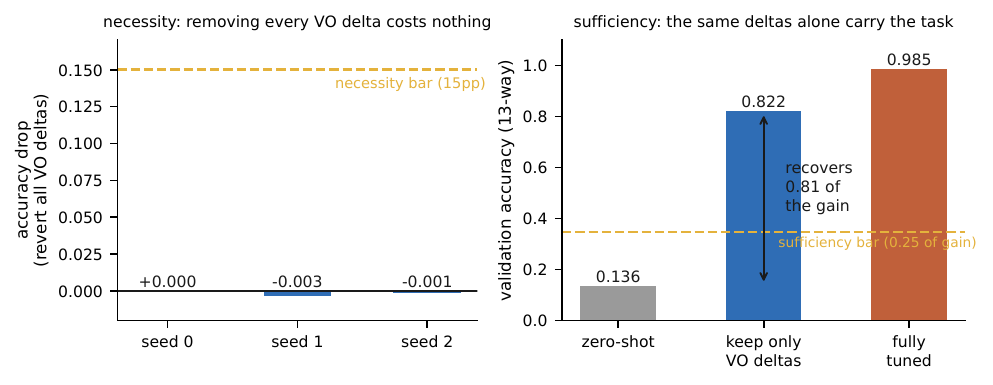}
\caption{\textbf{A small accuracy-reversion effect coexists with substantial retention recovery.} Left: VO-revert accuracy costs on three chart tuning seeds, as reported in the original audit. Right: base, VO-retained, and full-model accuracies on the primary 780-example chart evaluation set. VO retention uses pretrained LayerNorms and recovers 0.808 of the gain. This is component-level recovery relative to the pretrained network.}
\label{fig:inversion}
\end{figure}

Three additional comparisons sharpen the interpretation. First, Charts and EuroSAT retain the reversion/retention contrast in cross-entropy, while all five settings lose correct-class margin after VO reversion. Second, retaining only VO weight updates outperforms ten equally sized random delta subsets in each setting, including unsaturated Aircraft. Third, the retained models correct overlapping sets of pretrained errors: joint correction ranges from 0.250 on Aircraft to 0.814 on EuroSAT. These comparisons establish component-specific recovery and quantify its metric and task dependence.

Semantic readouts also guide effective removal interventions. Attention heads with delta directions that pass our task-semantic readout criteria appear in all five setups. On Charts, replacing financial text anchors with generic anchors while holding images and training hyperparameters fixed reduces coherent heads from 41 to zero. Readout-guided removal within the VO-retained model produces larger recovery collapse than matched-random removal in all five settings. Component retention measures recoverable performance from the whole update group; readout-guided removal measures the behavioral effect of semantic selection within that group.

\noindent\textbf{Contributions.}
(i) We pair component reversion and retention across accuracy, loss, and margin to measure the functional role of fine-tuning updates (Sections~\ref{sec:redundancy}--\ref{sec:loss}). (ii) We test the specificity and overlap of retained recovery using per-layer count-matched random subsets and shared corrections of pretrained errors (Sections~\ref{sec:random}--\ref{sec:overlap}). (iii) We evaluate semantic delta readouts through anchor substitution and matched-random direction removal within VO-retained models (Sections~\ref{sec:directions}--\ref{sec:alignment}). Together, these experiments distinguish task-aligned structure, removal cost, and recoverable performance.

\section{Related Work}
\label{sec:related}
\noindent\textbf{Semantic weight reading.}
The attention VO decomposition \citep{elhage2021mathematical} supports activation-based text decompositions of CLIP \citep{gandelsman2024textspan} and the weight-based approach of SITH \citep{gentile2026sith}. The latter reports semantic alignment in fine-tuning deltas despite limited rotation of the absolute weights. Concept-activation vectors \citep{kim2018tcav}, linear probes \citep{alain2018probes}, and sparse feature methods \citep{elhage2022superposition,cunningham2023sae,bricken2023monosemanticity,templeton2024scaling} provide related ways to describe learned representations. Our focus is the relationship between a semantic description of a weight delta and the behavior of edited checkpoints. We address this relationship at two explicit scales: component retention and readout-guided direction removal.

\noindent\textbf{Localization and intervention.}
Causal tracing and model editing \citep{meng2022rome,decao2021editing,mitchell2022mend,dai2022knowledge} test parameter or activation interventions. Hase et al.\ \citep{hase2023localization} show that localization by causal tracing does not necessarily predict editing success. Interpretability illusions \citep{bolukbasi2021illusion}, automated circuit discovery \citep{conmy2023acdc}, and sparse feature circuits \citep{marks2024sparsefeature} likewise motivate validating a readout against a specified intervention. Here the relevant distinction is between the effect of removing an update and the improvement recovered by retaining it.

\noindent\textbf{Redundancy and self-repair.}
Backup heads in the indirect object identification (IOI) circuit \citep{wang2023ioi} and the Hydra effect \citep{mcgrath2023hydra,rushing2024selfrepair} demonstrate that ablation effects depend on surrounding computation. Our experiments operate on fine-tuned weights and compare component reversion with component retention. After a weight edit, downstream computation remains active and can compensate during the forward pass. Our measurements characterize the behavior recovered by retained updates in this shared network.

\noindent\textbf{Sparse adaptation and merging.}
Task arithmetic \citep{ilharco2023task}, the model-merging methods TIES \citep{yadav2023ties}, DARE \citep{yu2024dare}, and intrinsic-dimensionality results \citep{aghajanyan2021intrinsic} show that useful adaptation can survive substantial parameter restriction. Grafting \citep{panigrahi2023grafting}, TaLoS \citep{iurada2025talos}, and Localize-and-Stitch \citep{he2024localize} identify task-relevant delta subsets. Singular-vector interventions also inform analyses of fine-tuned language models \citep{jin2025rlheals}. We connect semantic VO readouts to architectural-component retention, and test whether an equally sized random subset accounts for the observed recovery. Strong MLP retention is also compatible with the role of feed-forward layers as writable memories \citep{geva2021kv}.

\section{Reading and Testing Weight Deltas}
\label{sec:prelim}
\noindent\textbf{Setting.}
Let $\theta_0$ and $\theta_{\mathrm{ft}}$ denote the pretrained and tuned parameters, with $\Delta\theta=\theta_{\mathrm{ft}}-\theta_0$. Classification uses normalized image embeddings and frozen class-text anchors. We audit the final six blocks of CLIP ViT-L/14 ($w=1024$, 16 heads per block, 96 audited heads) and ViT-B/16 (72 audited heads). The text tower, final visual layer normalization (LayerNorm), visual projection, and logit scale remain fixed. The reading instrument follows the per-head effective operator
\begin{equation}
\label{eq:vo}
\mathrm{VO}_h^{(\ell)}(\theta)=W_O^{(\ell,h)}W_V^{(\ell,h)}\in\mathbb{R}^{w\times w}.
\end{equation}
A residual direction $u$ is mapped to the joint embedding through the frozen gain $\gamma$ and visual projection $P$:
\begin{equation}
\label{eq:proj}
e(u)=\frac{(u\odot\gamma)P}{\|(u\odot\gamma)P\|}.
\end{equation}
The dictionary $\mathcal A$ contains text embeddings, called atoms: 1,076 generic atoms, 21 competitors representing alternative senses of ambiguous phrases, and domain probes $\mathcal F$. Orthogonal matching pursuit (OMP) greedily selects $M=10$ atoms to approximate a mapped direction. A direction is domain-coherent when it satisfies
\begin{equation}
\label{eq:double}
\mathrm{OMP}_{M=10}(e;\mathcal A)\cap\mathcal F\ne\emptyset
\quad\wedge\quad
\max_{a\in\mathcal F}|\langle e,a\rangle|>\tau_{0.999},
\end{equation}
where $\tau_{0.999}$ is the empirical 99.9th percentile of the same maximum alignment over 5,000 random unit directions, recomputed per embedding dimension. The first condition requires a domain atom to participate in the greedy decomposition in competition with generic and confound-specific atoms.

\begin{figure}[t]
\centering
\includegraphics[width=\textwidth]{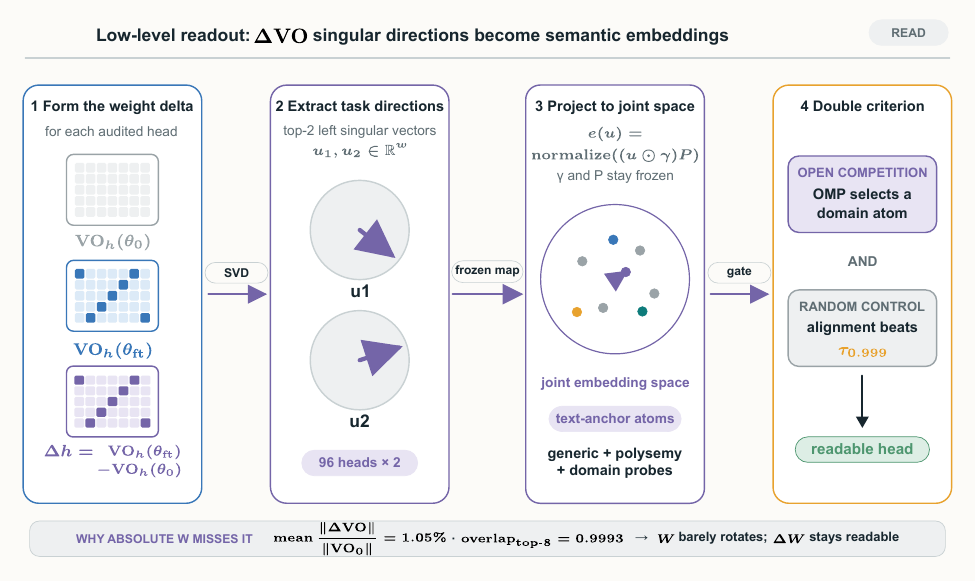}
\caption{\textbf{From effective VO deltas to semantic readouts.} The instrument forms the per-head effective delta, extracts its leading left singular vectors, maps them through the frozen gain and projection, and applies dictionary-selection and random-direction criteria. The top-eight pre/post subspace overlap of 0.9993 on Charts explains why delta reading reveals structure that changes little in the dominant absolute-weight subspace.}
\label{fig:readout}
\end{figure}
Singular value decomposition (SVD) extracts the two leading left singular vectors $u_h^{(1)}$ and $u_h^{(2)}$ from each effective delta. The readout acts on
\begin{equation}
\label{eq:delta}
\Delta_h=\mathrm{VO}_h(\theta_{\mathrm{ft}})-\mathrm{VO}_h(\theta_0),
\qquad \mathcal U_h=\{u_h^{(1)},u_h^{(2)}\}.
\end{equation}
A head is coherent if either mapped direction passes Eq.~\ref{eq:double}. The readout records the head, singular-rank index, and best-matching concept.

\noindent\textbf{Direction intervention.}
For a selected unit direction $u$, output-projection surgery is
\begin{equation}
\label{eq:surgery}
W_O^{(h)}\leftarrow(I-uu^\top)W_O^{(h)}.
\end{equation}
This acts on the full effective operator, including pretrained mass. The original chart audit records a mean ratio $\|u^\top\mathrm{VO}_{\mathrm{ft}}\|/\|u^\top\Delta_h\|=28.4$. In that full-model audit, all selected directions are precomputed from the original delta and then projected out. In the revision-stage VO-retained experiment, the multi-hit procedure instead recomputes the delta SVD on the current edited head at each selected rank; subsequent edits in a head are sequential readout-guided interventions. Each experiment uses matched-random controls with the same head slots and numbers of projections. The VO-retained direction experiment retains the first LayerNorm update, following its original condition definition; the component-retention headline experiments below use pretrained LayerNorms.

\noindent\textbf{Component interventions.}
We distinguish VO, query--key (QK), MLP, and two LayerNorm update groups. QK comprises the attention query and key projections; LN$_1$ and LN$_2$ denote the layer normalizations preceding attention and the MLP, respectively. A reversion starts from the tuned network and restores the selected parameters to their pretrained values. A retention starts from the tuned network and restores the complementary updates. Bias handling is specified explicitly in Appendix~\ref{app:instruments}: the VO revert restores $V$ biases but leaves the output-projection bias tuned, whereas standard VO retention includes both biases. The symbols $V_C$ and $K_C$ denote reversion and retention with these respective bias conventions. Weights-only VO retention, used for random-subset comparisons, restores all biases and both LayerNorms.
\begin{equation}
\label{eq:recovery}
\mathrm{Nec}_A(C)=A(\theta_{\mathrm{ft}})-A(V_C),\qquad
\mathrm{Rec}_A(C)=\frac{A(K_C)-A(\theta_0)}{A(\theta_{\mathrm{ft}})-A(\theta_0)}.
\end{equation}
Recovery measures sufficiency relative to a shared pretrained network. The frozen parts of that network remain active in every retention condition.

\noindent\textbf{Operational redundancy.}
For two distinct retained update groups, we summarize substantial recovery with low reversion costs by
\begin{equation}
\label{eq:redundant}
\min\{\mathrm{Rec}_A(C_1),\mathrm{Rec}_A(C_2)\}\ge\tau,\qquad
\max\{\mathrm{Nec}_A(C_1),\mathrm{Nec}_A(C_2)\}\le\epsilon.
\end{equation}
We use the point-estimate threshold $\epsilon=0.05$ for low accuracy necessity. Appendix~\ref{app:protocol} documents its original use in the VO dissociation test and its subsequent extension to all component groups. For VO and MLP, the paired minimum recoveries are 0.808, 0.547, 0.869, 0.428, and 0.438. The five-point criterion holds for the first three settings. Per-example overlap additionally measures the extent to which these retentions recover the same behavior. We define operational redundancy at the level of recovered behavior under these interventions.

\noindent\textbf{Continuous metrics and uncertainty.}
For cross-entropy $L$ and mean correct-class margin $M=z_y-\max_{k\ne y}z_k$, define
\begin{align}
N_L(C)&=L(V_C)-L(\theta_{\mathrm{ft}}),&
R_L(C)&=\frac{L(\theta_0)-L(K_C)}{L(\theta_0)-L(\theta_{\mathrm{ft}})},\\
N_M(C)&=M(\theta_{\mathrm{ft}})-M(V_C),&
R_M(C)&=\frac{M(K_C)-M(\theta_0)}{M(\theta_{\mathrm{ft}})-M(\theta_0)}.
\end{align}
All five evaluated gain denominators are positive. Paired bootstrap intervals use 2,000 example resamples, with all compared conditions evaluated on the same sampled indices. They quantify evaluation-example uncertainty conditional on the checkpoints. Random-mask ranges describe variation across mask seeds separately.

\section{Experimental Design}
\label{sec:protocol}
The design addresses three linked questions: whether small reversion costs conceal retained recovery, whether that recovery depends on the selected component and overlaps across examples, and whether semantic selection identifies an effective direction-removal procedure. Each comparison below targets one of these questions on fixed checkpoints and evaluation populations.

\noindent\textbf{Controlled semantic writing.}
Charts contain 13 financial concepts, with 400 training images and 60 evaluation images per concept. Fine-tuning uses cross-entropy against frozen chart-caption anchors, learning rate $10^{-5}$, batch size 48, and three epochs. The trainable surface is the final six visual blocks; the final visual LayerNorm, projection, and text tower remain frozen. Chart anchors overlap the domain probe vocabulary, so this is a controlled test of whether the recipe can write readable directions. A null tune uses identical images and hyperparameters with generic text anchors, isolating the effect of anchor choice under this recipe.

\noindent\textbf{Standard datasets.}
Flowers102, EuroSAT, and Aircraft use the same adaptation surface and five epochs, with 102, 10, and 100 frozen class-text anchors respectively. Flowers102 uses its training and validation splits for training and its test split for evaluation. Aircraft uses trainval and test. EuroSAT uses a seeded 80/20 split. Evaluation uses 2,000 test examples drawn by uniform random sampling with a fixed seed; per-class counts are supplied with the data. The ViT-B/16 condition repeats Flowers102. Full split and checkpoint details appear in Appendix~\ref{app:instruments}.

\noindent\textbf{Matched random retention.}
For each audited layer of width $w$, a random mask retains $2w^2$ scalar entries of the actual delta from the $V$, $O$, $Q$, $K$, and two MLP weight matrices. This exactly matches the number of VO weight entries in that layer. All other updates, biases, and LayerNorm parameters are reverted. Ten fixed seeds (7000--7009) are evaluated without rescaling or retraining. Matching controls parameter count and layer allocation; the retained delta norm is allowed to vary. We compare against VO weight retention with the same bias and LayerNorm treatment.

\noindent\textbf{Example-level comparisons.}
We compare VO retention, MLP retention, and the full model using prediction agreement on all evaluation examples and on the fixed set of errors of the pretrained model. Within that error set, we report the fraction corrected by both retained models and its excess over the product of their marginal correction rates. Shared correct answers distinguish recovered behavior from agreement on an incorrect label.

\noindent\textbf{Protocol history.}
The original audit hypothesized a strong dissociation: low VO necessity together with VO recovery at most 0.25 and MLP recovery at least 0.50. All five VO recoveries exceed 0.25, so that hypothesis fails. The empirical finding is instead substantial component retention with setting-dependent necessity and overlap. Appendix~\ref{app:protocol} records the decision criteria and separates original criteria from revision-stage analyses. The internal pre-specified protocol and its timestamp status are described there.

\section{Results}
\label{sec:results}
The results address three questions in turn: how much adaptation each component recovers, how that recovery compares with random subsets and overlaps across examples, and how semantic readouts guide behavioral intervention.
\subsection{Low Accuracy Necessity Coexists with Substantial Recovery}
\label{sec:redundancy}
\begin{table}[htbp]
\centering
\caption{Accuracy and recovery relative to the shared pretrained network. The first four settings use ViT-L/14; the final setting uses ViT-B/16. Both retained models use pretrained LayerNorm parameters. Bias conventions are specified in Appendix~\ref{app:instruments}.}
\label{tab:matrix}
\small
\setlength{\tabcolsep}{4pt}
\begin{tabular}{@{}lccccc@{}}
\toprule
Setting & $n$ & Base & Full & $\mathrm{Rec}_A(\mathrm{VO})$ & $\mathrm{Rec}_A(\mathrm{MLP})$ \\
\midrule
Charts & 780 & 0.1359 & 0.9846 & 0.808 & 0.989 \\
Flowers102 & 2000 & 0.7870 & 0.9900 & 0.547 & 0.862 \\
EuroSAT & 2000 & 0.4695 & 0.9770 & 0.869 & 0.960 \\
Aircraft & 2000 & 0.3105 & 0.7430 & 0.428 & 0.682 \\
Flowers102 (B/16) & 2000 & 0.6970 & 0.9735 & 0.438 & 0.808 \\
\bottomrule
\end{tabular}
\end{table}

\begin{table}[htbp]
\centering
\caption{Accuracy necessity costs with paired-bootstrap 95\% intervals. Values are accuracy fractions: 0.05 is five accuracy points. Reversion operators follow the bias convention in Appendix~\ref{app:instruments}; VO reversion leaves the output-projection bias tuned.}
\label{tab:necessity}
\small
\setlength{\tabcolsep}{4pt}
\begin{tabular}{@{}lccc@{}}
\toprule
Setting & VO revert & QK revert & MLP revert \\
\midrule
Charts & 0.000 [-0.008, 0.008] & 0.012 [0.004, 0.021] & 0.014 [0.001, 0.027] \\
Flowers102 & 0.015 [0.009, 0.022] & 0.003 [-0.001, 0.006] & 0.050 [0.040, 0.060] \\
EuroSAT & 0.010 [0.005, 0.016] & 0.004 [0.001, 0.008] & 0.024 [0.016, 0.032] \\
Aircraft & 0.061 [0.046, 0.076] & 0.036 [0.024, 0.049] & 0.127 [0.108, 0.148] \\
Flowers102 (B/16) & 0.030 [0.022, 0.039] & 0.005 [0.001, 0.009] & 0.124 [0.109, 0.139] \\
\bottomrule
\end{tabular}
\end{table}

VO-only recovery ranges from 0.428 to 0.869; MLP-only recovery ranges from 0.682 to 0.989 (Table~\ref{tab:matrix}). MLP retention recovers more of the accuracy gain in every setting. The stronger recoveries, such as Charts and EuroSAT, approach the full-model improvement; values around 0.43--0.55 represent substantial partial recovery. QK-only recovery is lower, ranging from 0.132 to 0.354 (Appendix~\ref{app:continuous}).

\label{sec:inversion}
Charts give a direct example of the distinction. The VO-retained model scores 0.8218, compared with 0.1359 pretrained and 0.9846 fully tuned. VO reversion leaves accuracy at 0.9846. MLP reversion costs 0.0141, while MLP retention recovers 0.9894 of the gain. Thus low VO accuracy necessity coexists with substantial recovery from VO updates, and both component costs satisfy the five-point criterion.

Table~\ref{tab:necessity} completes all fifteen component-setting cells. Twelve have point estimates at most 0.05. Flowers MLP is 0.0495, rounding to 0.050, with an interval spanning the threshold. Aircraft VO costs 0.0605, while Aircraft and ViT-B/16 MLP reverts cost 0.1270 and 0.1240. The low-necessity criterion consequently holds for all tested groups in Charts, Flowers102 on ViT-L/14, and EuroSAT. Aircraft and ViT-B/16 exhibit substantial retained recovery alongside larger component costs.

Independent LayerNorm controls recover 0.00 of the accuracy gain when retaining LN$_1$, LN$_2$, or both. Directly evaluated VO retention without LN$_1$ and MLP retention without LN$_2$ match their corresponding norm-retaining conditions to the reported precision. The headline results report these directly evaluated component retentions with pretrained LayerNorms.

\subsection{The Contrast Extends to Loss, While Margins Reveal Additional Contributions}
\label{sec:loss}
\begin{table}[htbp]
\centering
\caption{Cross-entropy necessity and retention recovery. The VO loss-cost interval spans zero on Charts and EuroSAT, while VO retention recovers 74.6\% and 82.2\% of the loss improvement. Table~\ref{tab:continuous} reports all three components and margins.}
\label{tab:loss}
\small
\setlength{\tabcolsep}{4pt}
\begin{tabular}{@{}lccc@{}}
\toprule
Setting & $N_L(\mathrm{VO})$ [95\% CI] & $R_L(\mathrm{VO})$ & $R_L(\mathrm{MLP})$ \\
\midrule
Charts & -0.0081 [-0.0245, 0.0055] & 0.746 & 0.988 \\
Flowers102 & 0.0395 [0.0254, 0.0531] & 0.631 & 0.880 \\
EuroSAT & 0.0035 [-0.0154, 0.0211] & 0.822 & 0.980 \\
Aircraft & 0.2024 [0.1710, 0.2337] & 0.505 & 0.698 \\
Flowers102 (B/16) & 0.1146 [0.0954, 0.1357] & 0.486 & 0.846 \\
\bottomrule
\end{tabular}
\end{table}

Charts and EuroSAT retain the reversion/retention contrast in cross-entropy (Table~\ref{tab:loss}). VO-revert loss costs are $-0.0081$ and $0.0035$, with intervals spanning zero, whereas VO-only retention recovers 0.746 and 0.822 of the loss improvement. The contrast therefore extends beyond thresholded accuracy. Across the five settings, VO-only loss recovery is 0.486--0.822 and MLP-only loss recovery is 0.698--0.988.

Loss necessity varies by setting. The VO costs on Flowers, Aircraft, and ViT-B/16 are 0.0395, 0.2024, and 0.1146, each with a positive interval. Aircraft combines the lowest full-model accuracy (0.7430) with appreciable loss necessity and partial retention recovery. These results show that the measured contribution depends on both the task and the evaluation metric.

Correct-class margins make this distinction explicit. VO reversion lowers mean margin by 1.519, 1.458, 2.940, 0.981, and 1.363 in setting order, with positive intervals in all five cases. Charts accuracy stays at 98.46\%, but its mean margin falls from 7.576 to 6.057. VO-only margin recovery is 0.225--0.283, compared with 0.470--0.687 for MLP. Recovery is therefore stronger for classification accuracy than for correct-class margin. Appendix~\ref{app:continuous} reports loss and margin necessity and recovery for VO, QK, and MLP.

\subsection{Recovery Is Component-Specific under Count Matching}
\label{sec:random}
\begin{table}[htbp]
\centering
\caption{Loss recovery under per-layer parameter-count matching. Both weights-only VO and random masks use pretrained biases and LayerNorms. Ranges describe ten mask seeds, not confidence intervals.}
\label{tab:random}
\small
\setlength{\tabcolsep}{4pt}
\begin{tabular}{@{}lccc@{}}
\toprule
Setting & Weights-only VO & Random mean & Random range \\
\midrule
Charts & 0.745 & 0.555 & [0.554, 0.556] \\
Flowers102 & 0.629 & 0.344 & [0.344, 0.345] \\
EuroSAT & 0.820 & 0.609 & [0.608, 0.611] \\
Aircraft & 0.504 & 0.280 & [0.279, 0.281] \\
Flowers102 (B/16) & 0.484 & 0.260 & [0.259, 0.261] \\
\bottomrule
\end{tabular}
\end{table}

The separate weights-only VO condition exceeds every one of the ten sampled random masks in loss recovery in every setting (Table~\ref{tab:random}). The advantage over the mean random mask is approximately 0.19--0.29 of the loss gain. It persists on unsaturated Aircraft: weights-only VO accuracy is 0.4930, versus at most 0.4035 across random masks, with full-model accuracy 0.7430.

Random subsets also recover part of the improvement, consistent with distributed adaptation. VO retention achieves greater recovery with the same parameter count and layer allocation. Its advantage shows that the choice of retained weights matters under this control. Retained delta norms vary across subsets, as specified in Section~\ref{sec:protocol}.

Bias effects are small but measurable. Charts accuracy is 0.8218 under standard VO retention and 0.8192 with all biases reverted; EuroSAT changes from 0.9105 to 0.9080. Table~\ref{tab:random} uses the directly measured weights-only results to match the bias treatment of the random controls.

\subsection{Shared Corrections Quantify Operational Redundancy}
\label{sec:overlap}
\begin{table}[htbp]
\centering
\caption{Prediction agreement on all evaluation examples. Shared corrections of pretrained errors are reported separately to distinguish overlap in recovered behavior from shared mistakes.}
\label{tab:agreement}
\small
\setlength{\tabcolsep}{4pt}
\begin{tabular}{@{}lccc@{}}
\toprule
Setting & VO/full & MLP/full & VO/MLP \\
\midrule
Charts & 0.8256 & 0.9859 & 0.8321 \\
Flowers102 & 0.9015 & 0.9670 & 0.9290 \\
EuroSAT & 0.9190 & 0.9700 & 0.9255 \\
Aircraft & 0.5500 & 0.7090 & 0.6820 \\
Flowers102 (B/16) & 0.8315 & 0.9375 & 0.8725 \\
\bottomrule
\end{tabular}
\end{table}

Prediction agreement between VO retention and the full model ranges from 0.5500 to 0.9190; agreement between MLP retention and the full model ranges from 0.7090 to 0.9859 (Table~\ref{tab:agreement}). VO/MLP agreement ranges from 0.6820 to 0.9290. These quantities describe shared predictions, including shared mistakes.

On the pretrained model's fixed error set, the fractions corrected by both retentions are 0.7893, 0.5211, 0.8143, 0.2502, and 0.3977. Joint correction exceeds the product of the marginal correction rates by 0.018, 0.084, 0.045, 0.114, and 0.096; the corresponding intervals are $[0.009,0.027]$, $[0.068,0.100]$, $[0.034,0.056]$, $[0.104,0.124]$, and $[0.083,0.109]$. This comparison characterizes association in corrected examples; example difficulty can also contribute to that association.

Overlap is strongest on Charts and EuroSAT and weaker on Aircraft. Together with the retention results, it supports an operational interpretation in which different updates recover overlapping behavior through the shared pretrained network. The measured overlap is between the outputs of two retention conditions applied to serial, jointly trained attention and MLP blocks.

\subsection{Readout-Guided Removal Reduces Recovery More Than Random Removal}
\label{sec:directions}
Starting from the norm-retaining VO-only condition, we apply the readout-guided surgery of Eq.~\ref{eq:surgery} and compare its recovery collapse with same-head, same-count random projections. The identified-minus-random collapse differences are positive in all five settings (Table~\ref{tab:direction}). Bootstrap intervals are conditional on the three sampled controls; random-control collapse ranges are at most 0.009.

\begin{table}[tbp]
\centering
\caption{Readout-guided minus random collapse in normalized accuracy recovery. The intervention applies sequential removal projections within VO-retained models. Ranges describe the three random seeds.}
\label{tab:direction}
\small
\begin{tabular}{@{}lccc@{}}
\toprule
Setting & Collapse difference [95\% CI] & Random range & Majority criterion\\
\midrule
Charts & 0.366 [0.325, 0.409] & 0.004 & Fail\\
Flowers102 & 0.081 [0.039, 0.124] & 0.005 & Fail\\
EuroSAT & 0.736 [0.699, 0.775] & 0.002 & Pass\\
Aircraft & 0.102 [0.068, 0.136] & 0.009 & Fail\\
Flowers102 (B/16) & 0.177 [0.138, 0.216] & 0.007 & Fail\\
\bottomrule
\end{tabular}
\end{table}
The specified majority-contribution criterion required removal to eliminate at least half of VO-retained recovery and at least twice the random collapse. Only EuroSAT meets it. Charts loses about 0.37 of normalized recovery, below its required 0.405. We report the intervention as a control-separated contribution of the readout-guided procedure, with majority collapse in one setting.

We use removal because it can be implemented directly through the output-projection weights. By contrast, retaining only selected rank-one delta components generally combines pretrained and tuned value paths, whose sum can exceed the original per-head rank bottleneck. The reported effect therefore measures sequential readout-guided removal, including its action on pretrained mass. Component retention separately measures recovery from an entire update group.

\subsection{Semantic Alignment and Its Attribution}
\label{sec:alignment}
\begin{figure}[tbp]
\centering
\includegraphics[width=\textwidth]{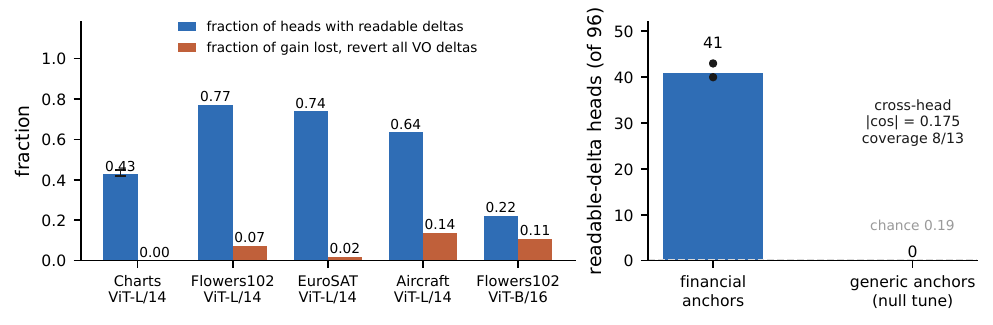}
\caption{\textbf{Semantic alignment and accuracy necessity measure different properties.} Left: coherent-head fraction and the fraction of tuning accuracy gain lost after VO reversion. Right: Charts produces 41 coherent heads under financial anchors and zero under generic anchors; points show 40, 41, and 43 heads across the original three tuning seeds. The null-tune comparison holds images and hyperparameters fixed.}
\label{fig:alignment}
\end{figure}
Coherent delta heads number 41/96 on Charts, 74/96 on Flowers102, 71/96 on EuroSAT, 61/96 on Aircraft, and 16/72 on ViT-B/16. On Charts, counts range from 40 to 43 across three tuning seeds. The coherent directions have mean cross-head absolute cosine 0.175 and best matches covering eight of thirteen trained concepts. These measurements show alignment spread across heads and concepts.

The chart null tune converges from loss 0.527 to 0.037 but yields zero coherent heads. Changing the text anchors thus changes the measured semantic structure under fixed images and hyperparameters. The matched comparison isolates the effect of anchor choice while preserving image supervision.

Relative chart deltas are small: mean $\|\Delta\mathrm{VO}\|/\|\mathrm{VO}_0\|=0.0105$, with top-eight pre/post singular-subspace overlap 0.9993. The MLP-to-VO relative-write ratio is 1.03, indicating comparable relative update magnitudes. An additional ViT-B/32 chart run yields six coherent heads of 72, a zero-head null tune, and a 2.9-point VO-revert cost. Its coherent-head fraction fails the original 0.20 replication bar. Because the random-direction threshold changes with embedding dimension, fractions are interpreted within architecture.

\subsection{Pretrained Readouts and Polysemy Controls}
\label{sec:pretrained}
The pretrained chart audit asks whether semantic hits reflect the financial sense of a phrase. A naive run identifies three coherent heads, including color and anatomy matches for candlestick and head-and-shoulders phrases. Adding 21 explicit competitors and using 41 chart-anchored probes reduces selected vectors from 59 to 37 and vectors passing both criteria from four to one. The remaining hit selects a person's head-and-shoulders portrait before the financial atom. The surviving hit therefore remains ambiguous between the portrait and financial senses.

Under the specified dictionary, the leading VO directions in the evaluated OpenAI checkpoints provide little unambiguous evidence of the target financial concepts. Explicit competitors make the semantic ambiguity visible in the decomposition. Because greedy selection depends on the candidate dictionary, we evaluate each dictionary condition separately.

OpenAI ViT-B/32 yields zero of 72 coherent pretrained heads. OpenCLIP ViT-L/14 trained on LAION-2B \citep{cherti2023openclip,schuhmann2022laion} yields four of 96, exceeding the audit's bound of two. These counts show variation across pretrained checkpoints. The semantic interpretation of the LAION hits remains unresolved because their detailed competitor-selection traces are unavailable (Figure~\ref{fig:pretrained}).

\section{Discussion}
\label{sec:discussion}
\noindent\textbf{Interpreting a small ablation effect.}
Necessity measures the effect of a particular edit in its surrounding network. It remains informative about that intervention even when retained recovery is high. The error is the further inference that a small accuracy cost establishes absence of functional contribution. Charts demonstrates the distinction in accuracy and loss, while its margin reduction shows why specifying the metric matters.

\noindent\textbf{What the controls distinguish.}
Random retention shows that the observed recovery is sensitive to which updates are retained, even under equal per-layer counts. Example-level co-correction quantifies overlap in recovered behavior. Together, these controls show that structured update groups recover shared correct answers through the pretrained residual network. Attention and MLP remain serial, jointly trained components of that network.

\noindent\textbf{Using semantic readouts.}
The readout identifies task-aligned weight structure and provides a target for controlled intervention. Anchor substitution strengthens its attribution in the chart setting. Component retention establishes recoverable performance on a broader substrate, and readout-guided removal connects semantic selection to behavior at a finer scale. The distinction between these substrates should remain explicit in editing, activation steering \citep{turner2023actadd}, and capability-removal studies.

\noindent\textbf{A practical evaluation sequence.}
Specify the parameter and bias mask, evaluate base and tuned models on identical examples, and pair each reversion with a clearly defined retention. Report raw performance, normalized recovery, and a continuous metric. Compare structured retention with matched random subsets and characterize shared corrections on pretrained errors. For direction interventions, state the projection operator, whether directions are fixed or recomputed, and how pretrained mass is affected. Retention and shared corrections directly measure recoverable performance and its overlap across update groups.

\section{Limitations}
\label{sec:limitations}
The evidence covers five CLIP classification setups under a shared recipe that fine-tunes the final six visual blocks with frozen text anchors and readout. The observed recovery and overlap characterize this model family and recipe. Learning-rate changes, broader trainable surfaces, other model families, and per-layer retention would test how recoverable adaptation is distributed under different training conditions. Restricting adaptation to VO with a method such as low-rank adaptation (LoRA) \citep{hu2022lora} changes the available update groups and remains an untested extension.

Operational redundancy is task- and metric-dependent. Charts and EuroSAT show low VO necessity alongside substantial retention in both accuracy and cross-entropy; margin reductions reveal contributions in every setting, and Aircraft and ViT-B/16 have larger component-reversion costs. The paired-bootstrap intervals measure evaluation-example variation conditional on each checkpoint. The random-retention comparison controls parameter count within each layer; it does not match delta norm.

Component retention measures recovery through the shared pretrained network, where attention and MLP updates were trained jointly and act in series. Shared corrections quantify overlapping behavior rather than identical internal computation. Readout-guided direction removal measures a finer causal contribution: the implemented sequential projections can also remove pretrained mass, and they do not constitute direction-only retention. These operator distinctions determine the scope of the corresponding conclusions.

\section{Conclusion}
\label{sec:conclusion}
A small accuracy drop after reversion can coexist with substantial recovery when the same component's updates are retained. In the evaluated CLIP setups, structured VO retention outperforms count-matched random subsets, and VO and MLP retentions correct overlapping pretrained errors. Readout-guided removal further links semantic selection to changes in recovered behavior. Loss and margin measurements show how the apparent contribution changes with the metric. These findings support evaluating readable updates through paired removal and retention, with explicit controls for the selected parameters and intervention. Operational redundancy describes the overlapping recovery observed under this CLIP fine-tuning recipe.

\subsubsection*{Broader Impact Statement}
This work improves the interpretation of ablation-based model audits. A capability may persist after selected updates are removed, and a small measured accuracy change does not establish that those updates carried no useful adaptation. Audits of capability removal should evaluate the edited model's behavior and continuous scores directly, alongside retention controls where appropriate.

\subsubsection*{Reproducibility Statement}
The accompanying reproduction package includes the chart generator, training and intervention scripts, environment specifications, saved result files, and per-example outputs. A CPU workflow rebuilds result summaries and recomputes prediction agreement; the manuscript source package additionally regenerates its component tables. Appendix~\ref{app:reproducibility} gives the commands and their coverage. Checkpoint hashes are supplied; checkpoint weights are not included in the archive. Model-level reruns require the specified datasets and pretrained weights, followed by training or provision of the corresponding fine-tuned checkpoints.

\subsubsection*{AI Assistance Statement}
We used AI tools for language and drafting assistance. The experimental design, choice of controls, analyses, and scientific judgments are the authors' own.

\bibliography{refs}

@inproceedings{gentile2026sith,
  title={From Weights to Concepts: Data-Free Interpretability of {CLIP} via Singular Vector Decomposition},
  author={Gentile, Francesco and Dall'Asen, Nicola and Tonini, Francesco and Mancini, Massimiliano and Vaquero, Lorenzo and Ricci, Elisa},
  booktitle={IEEE/CVF Conference on Computer Vision and Pattern Recognition (CVPR)},
  year={2026},
  note={arXiv:2603.24653}
}

@inproceedings{hase2023localization,
  title={Does Localization Inform Editing? {S}urprising Differences in Causality-Based Localization vs. Knowledge Editing in Language Models},
  author={Hase, Peter and Bansal, Mohit and Kim, Been and Ghandeharioun, Asma},
  booktitle={Advances in Neural Information Processing Systems (NeurIPS)},
  year={2023}
}

@article{mcgrath2023hydra,
  title={The Hydra Effect: Emergent Self-Repair in Language Model Computations},
  author={McGrath, Thomas and Rahtz, Matthew and Kram{\'a}r, J{\'a}nos and Mikulik, Vladimir and Legg, Shane},
  journal={arXiv preprint arXiv:2307.15771},
  year={2023}
}

@inproceedings{rushing2024selfrepair,
  title={Explorations of Self-Repair in Language Models},
  author={Rushing, Cody and Nanda, Neel},
  booktitle={International Conference on Machine Learning (ICML)},
  year={2024}
}

@inproceedings{wang2023ioi,
  title={Interpretability in the Wild: a Circuit for Indirect Object Identification in {GPT}-2 Small},
  author={Wang, Kevin and Variengien, Alexandre and Conmy, Arthur and Shlegeris, Buck and Steinhardt, Jacob},
  booktitle={International Conference on Learning Representations (ICLR)},
  year={2023}
}

@inproceedings{yadav2023ties,
  title={{TIES}-Merging: Resolving Interference When Merging Models},
  author={Yadav, Prateek and Tam, Derek and Choshen, Leshem and Raffel, Colin and Bansal, Mohit},
  booktitle={Advances in Neural Information Processing Systems (NeurIPS)},
  year={2023}
}

@inproceedings{yu2024dare,
  title={Language Models are Super {M}ario: Absorbing Abilities from Homologous Models as a Free Lunch},
  author={Yu, Le and Yu, Bowen and Yu, Haiyang and Huang, Fei and Li, Yongbin},
  booktitle={International Conference on Machine Learning (ICML)},
  year={2024}
}

@inproceedings{panigrahi2023grafting,
  title={Task-Specific Skill Localization in Fine-tuned Language Models},
  author={Panigrahi, Abhishek and Saunshi, Nikunj and Zhao, Haoyu and Arora, Sanjeev},
  booktitle={International Conference on Machine Learning (ICML)},
  year={2023}
}

@inproceedings{iurada2025talos,
  title={Efficient Model Editing with Task-Localized Sparse Fine-tuning},
  author={Iurada, Leonardo and Ciccone, Marco and Tommasi, Tatiana},
  booktitle={International Conference on Learning Representations (ICLR)},
  year={2025}
}

@article{he2024localize,
  title={Localize-and-Stitch: Efficient Model Merging via Sparse Task Arithmetic},
  author={He, Yifei and Hu, Yuzheng and Lin, Yong and Zhang, Tong and Zhao, Han},
  journal={Transactions on Machine Learning Research (TMLR)},
  year={2024}
}

@article{jin2025rlheals,
  title={{RL} Fine-Tuning Heals {OOD} Forgetting in {SFT}},
  author={Jin, Hangzhan and Luan, Sitao and Ni, Tianwei and Lyu, Sicheng and Rabusseau, Guillaume and Rabbany, Reihaneh and Precup, Doina and Hamdaqa, Mohammad},
  journal={arXiv preprint arXiv:2509.12235},
  year={2025}
}

@inproceedings{gargiulo2025tsv,
  title={Task Singular Vectors: Reducing Task Interference in Model Merging},
  author={Gargiulo, Antonio Andrea and Crisostomi, Donato and Bucarelli, Maria Sofia and Scardapane, Simone and Silvestri, Fabrizio and Rodol{\`a}, Emanuele},
  booktitle={IEEE/CVF Conference on Computer Vision and Pattern Recognition (CVPR)},
  year={2025}
}

@inproceedings{ilharco2023task,
  title={Editing Models with Task Arithmetic},
  author={Ilharco, Gabriel and Ribeiro, Marco Tulio and Wortsman, Mitchell and Gururangan, Suchin and Schmidt, Ludwig and Hajishirzi, Hannaneh and Farhadi, Ali},
  booktitle={International Conference on Learning Representations (ICLR)},
  year={2023}
}

@inproceedings{aghajanyan2021intrinsic,
  title={Intrinsic Dimensionality Explains the Effectiveness of Language Model Fine-Tuning},
  author={Aghajanyan, Armen and Gupta, Sonal and Zettlemoyer, Luke},
  booktitle={Annual Meeting of the Association for Computational Linguistics (ACL)},
  year={2021}
}

@article{bolukbasi2021illusion,
  title={An Interpretability Illusion for {BERT}},
  author={Bolukbasi, Tolga and Pearce, Adam and Yuan, Ann and Coenen, Andy and Reif, Emily and Vi{\'e}gas, Fernanda and Wattenberg, Martin},
  journal={arXiv preprint arXiv:2104.07143},
  year={2021}
}

@article{elhage2021mathematical,
  title={A Mathematical Framework for Transformer Circuits},
  author={Elhage, Nelson and Nanda, Neel and Olsson, Catherine and Henighan, Tom and Joseph, Nicholas and Mann, Ben and Askell, Amanda and Bai, Yuntao and Chen, Anna and Conerly, Tom and others},
  journal={Transformer Circuits Thread},
  year={2021},
  note={https://transformer-circuits.pub/2021/framework/index.html}
}

@inproceedings{gandelsman2024textspan,
  title={Interpreting {CLIP}'s Image Representation via Text-Based Decomposition},
  author={Gandelsman, Yossi and Efros, Alexei A. and Steinhardt, Jacob},
  booktitle={International Conference on Learning Representations (ICLR)},
  year={2024}
}

@inproceedings{radford2021clip,
  title={Learning Transferable Visual Models From Natural Language Supervision},
  author={Radford, Alec and Kim, Jong Wook and Hallacy, Chris and Ramesh, Aditya and Goh, Gabriel and Agarwal, Sandhini and Sastry, Girish and Askell, Amanda and Mishkin, Pamela and Clark, Jack and Krueger, Gretchen and Sutskever, Ilya},
  booktitle={International Conference on Machine Learning (ICML)},
  year={2021}
}

@inproceedings{cherti2023openclip,
  title={Reproducible Scaling Laws for Contrastive Language-Image Learning},
  author={Cherti, Mehdi and Beaumont, Romain and Wightman, Ross and Wortsman, Mitchell and Ilharco, Gabriel and Gordon, Cade and Schuhmann, Christoph and Schmidt, Ludwig and Jitsev, Jenia},
  booktitle={IEEE/CVF Conference on Computer Vision and Pattern Recognition (CVPR)},
  year={2023}
}

@inproceedings{schuhmann2022laion,
  title={{LAION}-5{B}: An Open Large-Scale Dataset for Training Next Generation Image-Text Models},
  author={Schuhmann, Christoph and Beaumont, Romain and Vencu, Richard and Gordon, Cade and Wightman, Ross and Cherti, Mehdi and Coombes, Theo and Katta, Aarush and Mullis, Clayton and Wortsman, Mitchell and others},
  booktitle={Advances in Neural Information Processing Systems (NeurIPS) Datasets and Benchmarks Track},
  year={2022}
}

@inproceedings{nilsback2008flowers,
  title={Automated Flower Classification over a Large Number of Classes},
  author={Nilsback, Maria-Elena and Zisserman, Andrew},
  booktitle={Indian Conference on Computer Vision, Graphics and Image Processing (ICVGIP)},
  year={2008}
}

@article{helber2019eurosat,
  title={{EuroSAT}: A Novel Dataset and Deep Learning Benchmark for Land Use and Land Cover Classification},
  author={Helber, Patrick and Bischke, Benjamin and Dengel, Andreas and Borth, Damian},
  journal={IEEE Journal of Selected Topics in Applied Earth Observations and Remote Sensing},
  year={2019}
}

@article{maji2013aircraft,
  title={Fine-Grained Visual Classification of Aircraft},
  author={Maji, Subhransu and Rahtu, Esa and Kannala, Juho and Blaschko, Matthew and Vedaldi, Andrea},
  journal={arXiv preprint arXiv:1306.5151},
  year={2013}
}

@inproceedings{meng2022rome,
  title={Locating and Editing Factual Associations in {GPT}},
  author={Meng, Kevin and Bau, David and Andonian, Alex and Belinkov, Yonatan},
  booktitle={Advances in Neural Information Processing Systems (NeurIPS)},
  year={2022}
}

@inproceedings{geva2021kv,
  title={Transformer Feed-Forward Layers Are Key-Value Memories},
  author={Geva, Mor and Schuster, Roei and Berant, Jonathan and Levy, Omer},
  booktitle={Conference on Empirical Methods in Natural Language Processing (EMNLP)},
  year={2021}
}

@inproceedings{hu2022lora,
  title={{LoRA}: Low-Rank Adaptation of Large Language Models},
  author={Hu, Edward J. and Shen, Yelong and Wallis, Phillip and Allen-Zhu, Zeyuan and Li, Yuanzhi and Wang, Shean and Wang, Lu and Chen, Weizhu},
  booktitle={International Conference on Learning Representations (ICLR)},
  year={2022}
}

@inproceedings{cunningham2023sae,
  title={Sparse Autoencoders Find Highly Interpretable Features in Language Models},
  author={Huben, Robert and Cunningham, Hoagy and Smith, Logan and Ewart, Aidan and Sharkey, Lee},
  booktitle={International Conference on Learning Representations (ICLR)},
  year={2024}
}

@article{bricken2023monosemanticity,
  title={Towards Monosemanticity: Decomposing Language Models with Dictionary Learning},
  author={Bricken, Trenton and Templeton, Adly and Batson, Joshua and Chen, Brian and Jermyn, Adam and Conerly, Tom and Turner, Nicholas L and Anil, Cem and Denison, Carson and Askell, Amanda and others},
  journal={Transformer Circuits Thread},
  year={2023}
}

@article{elhage2022superposition,
  title={Toy Models of Superposition},
  author={Elhage, Nelson and Hume, Tristan and Olsson, Catherine and Schiefer, Nicholas and Henighan, Tom and Kravec, Shauna and Hatfield-Dodds, Zac and Lasenby, Robert and Drain, Dawn and Chen, Carol and others},
  journal={Transformer Circuits Thread, arXiv:2209.10652},
  year={2022}
}

@inproceedings{kim2018tcav,
  title={Interpretability Beyond Feature Attribution: Quantitative Testing with Concept Activation Vectors ({TCAV})},
  author={Kim, Been and Wattenberg, Martin and Gilmer, Justin and Cai, Carrie and Wexler, James and Vi{\'e}gas, Fernanda and Sayres, Rory},
  booktitle={International Conference on Machine Learning (ICML)},
  year={2018}
}

@article{alain2018probes,
  title={Understanding Intermediate Layers Using Linear Classifier Probes},
  author={Alain, Guillaume and Bengio, Yoshua},
  journal={arXiv preprint arXiv:1610.01644v4},
  year={2018}
}

@inproceedings{conmy2023acdc,
  title={Towards Automated Circuit Discovery for Mechanistic Interpretability},
  author={Conmy, Arthur and Mavor-Parker, Augustine N and Lynch, Aengus and Heimersheim, Stefan and Garriga-Alonso, Adri{\`a}},
  booktitle={Advances in Neural Information Processing Systems (NeurIPS)},
  year={2023}
}

@inproceedings{marks2024sparsefeature,
  title={Sparse Feature Circuits: Discovering and Editing Interpretable Causal Graphs in Language Models},
  author={Marks, Samuel and Rager, Can and Michaud, Eric J and Belinkov, Yonatan and Bau, David and Mueller, Aaron},
  booktitle={International Conference on Learning Representations (ICLR)},
  year={2025}
}

@inproceedings{dai2022knowledge,
  title={Knowledge Neurons in Pretrained Transformers},
  author={Dai, Damai and Dong, Li and Hao, Yaru and Sui, Zhifang and Chang, Baobao and Wei, Furu},
  booktitle={Annual Meeting of the Association for Computational Linguistics (ACL)},
  year={2022}
}

@inproceedings{decao2021editing,
  title={Editing Factual Knowledge in Language Models},
  author={De Cao, Nicola and Aziz, Wilker and Titov, Ivan},
  booktitle={Conference on Empirical Methods in Natural Language Processing (EMNLP)},
  year={2021}
}

@inproceedings{mitchell2022mend,
  title={Fast Model Editing at Scale},
  author={Mitchell, Eric and Lin, Charles and Bosselut, Antoine and Finn, Chelsea and Manning, Christopher D},
  booktitle={International Conference on Learning Representations (ICLR)},
  year={2022}
}

@article{templeton2024scaling,
  title={Scaling Monosemanticity: Extracting Interpretable Features from {C}laude 3 {S}onnet},
  author={Templeton, Adly and Conerly, Tom and Marcus, Jonathan and Lindsey, Jack and Bricken, Trenton and Chen, Brian and Pearce, Adam and Citro, Craig and Ameisen, Emmanuel and Jermyn, Adam and others},
  journal={Transformer Circuits Thread},
  year={2024}
}

@article{turner2023actadd,
  title={Steering Language Models With Activation Engineering},
  author={Turner, Alexander Matt and Thiergart, Lisa and Leech, Gavin and Udell, David and Vazquez, Juan J and Mini, Ulisse and MacDiarmid, Monte},
  journal={arXiv preprint arXiv:2308.10248v5},
  year={2024}
}
\bibliographystyle{tmlr}
\appendix
\section{Instruments, Data, and Intervention Accounting}
\label{app:instruments}
\noindent\textbf{Architectures and training.}
The primary model is OpenAI CLIP ViT-L/14 with QuickGELU, width 1024, 24 layers, 16 heads per layer, head dimension 64, and embedding dimension 768. The final six visual blocks are trainable. ViT-B/16 and the additional ViT-B/32 chart control use the corresponding OpenAI checkpoints with QuickGELU. Fine-tuning uses learning rate $10^{-5}$ and batch size 48; Charts uses three epochs and the standard datasets use five. The text tower, final visual LayerNorm, visual projection, and logit scale remain fixed. Evaluations use the terminal checkpoints of the specified runs, subject to the original learning-validity gates, rather than selecting checkpoints by intervention outcomes.

\noindent\textbf{Evaluation populations.}
The chart training generator uses seed 0 and 400 images per concept. The training-time validation set uses seed 999 and 60 images per concept; the reported accuracy was 98.3\%. The intervention evaluation uses a separately rendered seed-4242 set of 780 images and gives 98.46\%, reported previously as 98.5\%. The two evaluation populations were rendered independently from the same generator. All paired comparisons in the new analysis use the seed-4242 set. A Wilson interval for its full-model accuracy is approximately $[0.973,0.991]$.

Flowers102 uses torchvision's train and validation splits for training and test for evaluation. Aircraft uses trainval and test. EuroSAT uses a seed-0 random permutation followed by an 80/20 train/test division. Each standard evaluation set is a seed-0 uniform random sample of 2,000 test examples. Label counts are supplied in the accompanying data file; Charts has exactly 60 examples per class. The fixed pretrained-error counts are 674, 426, 1,061, 1,379, and 606 in setting order. The analysis stores within-evaluation row indices, labels, predictions, losses, and margins; original dataset indices can be regenerated from the specified loader and seeds.

\noindent\textbf{Dictionary and sensitivity checks.}
The generic dictionary uses 1,076 ImageNet-derived and attribute atoms, embedded using averaged text templates and renormalized. Financial probes contain 41 chart-anchored phrases; 21 competitors name confounds such as wax candles and anatomical head-and-shoulders portraits. The top two left singular vectors of each effective VO delta are tested with ten greedy dictionary atoms and a 5,000-direction empirical null. The original instrument check on a strongly tuned small checkpoint detected 21 written heads and identified a deliberately smeared single-direction case. These checks assess instrument sensitivity for the selected construction.

\noindent\textbf{Exact bias conventions.}
VO reversion restores all $V$ rows and their input-projection biases, and all $O$ weight columns; the output-projection bias stays tuned. QK reversion restores $Q/K$ rows and their biases. MLP reversion restores both linear layers and both biases. All other updates, including LayerNorms, stay tuned in these reversion conditions. Standard VO retention keeps $V/O$ weights, $V$ biases, and the output-projection bias, while restoring QK, MLP, and both LayerNorms. Standard MLP retention keeps its two weight matrices and biases while restoring attention and both LayerNorms. QK retention keeps its weight rows and biases while restoring the remaining updates. Consequently the named VO reversion and retention operators differ in output-bias scope; the reported operational criteria refer to these explicit operators.

Weights-only VO retention additionally restores $V$ and output-projection biases. Every random mask starts with all eligible updates, biases, and LayerNorms restored, then adds back the selected scalar weight deltas. Within each layer, the fixed sampling order is $V,O,Q,K,\mathrm{MLP}_{\mathrm{in}},\mathrm{MLP}_{\mathrm{out}}$; a random permutation selects $2w^2$ of the $12w^2$ eligible entries. Seeds 7000--7009 determine the masks. The five weights-only VO loss recoveries are 0.745, 0.629, 0.820, 0.504, and 0.484.

\noindent\textbf{Direction-surgery implementation.}
The coherent hit list is selected from the full tuned model. For each hit in its stored order, the surgery routine recomputes the effective delta SVD on the current model, selects the stored rank, and projects that direction out of the corresponding $O$ columns. If a head has multiple hits, later directions can differ from its initial singular vectors. The direction experiment starts from VO retention with LN$_1$ kept; it is distinct from the no-LayerNorm retention in Table~\ref{tab:matrix}. Matched-random controls apply the same number and head allocation of sequential projections, using seeds 3000--3002. This intervention measures the readout-guided procedure as executed.

\section{Additional Results and Complete Continuous Metrics}
\label{app:extra}
\begin{figure}[htbp]
\centering
\includegraphics[width=\textwidth]{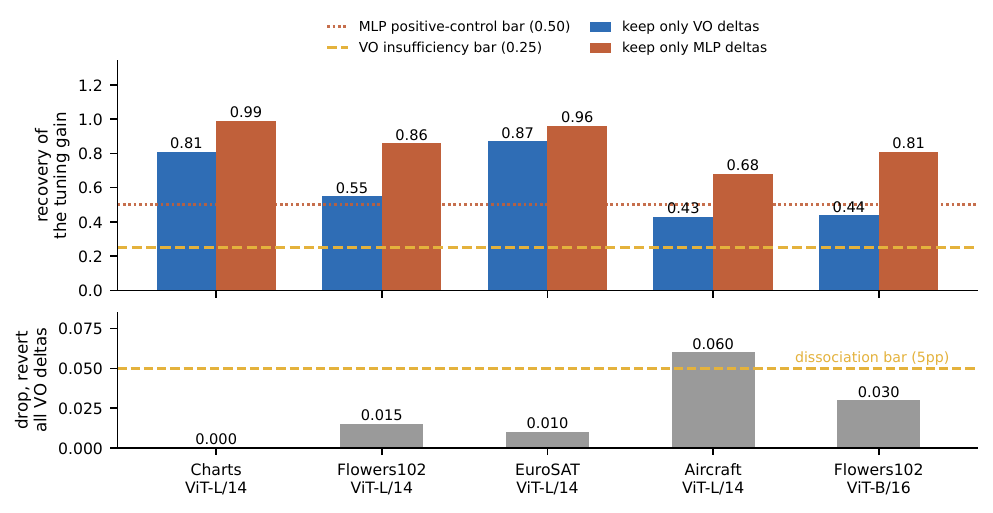}
\caption{\textbf{Component retention and VO necessity in the original audit.} Top: VO and MLP accuracy recoveries; the reference lines show the original VO insufficiency and MLP positive-control criteria. Bottom: VO-revert accuracy costs. The complete VO, QK, and MLP necessity measurements, including the larger MLP costs on Aircraft and ViT-B/16, appear in Table~\ref{tab:necessity}.}
\label{fig:redundancy}
\end{figure}
\begin{figure}[htbp]
\centering
\includegraphics[width=\textwidth]{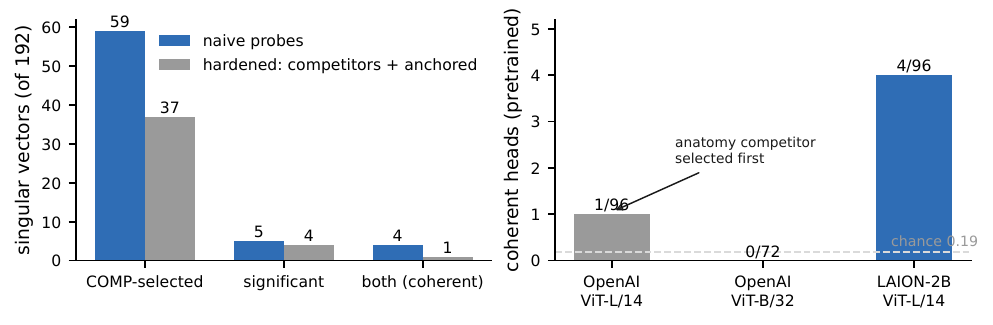}
\caption{\textbf{Pretrained semantic readouts depend on dictionary and corpus.} Left: selected, null-passing, and jointly passing vector counts under the original and competitor-enriched chart audits. Right: technically coherent pretrained heads across corpora and architectures. The sole OpenAI ViT-L/14 hit remains polysemy-confounded; LAION hits lack a full selection trace.}
\label{fig:pretrained}
\end{figure}
\begin{figure}[htbp]
\centering
\includegraphics[width=\textwidth]{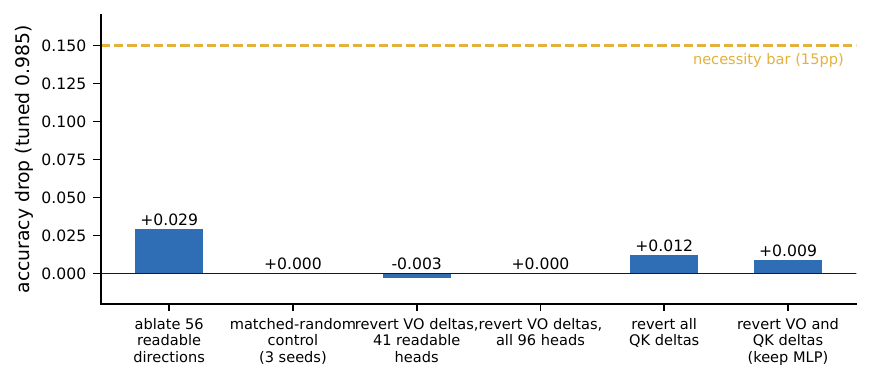}
\caption{\textbf{The original chart accuracy intervention battery.} Direction surgery and matched-random removal are shown alongside coherent-head VO, all-head VO, QK, and joint VO+QK reverts. The original 15-point line is a decision gate for this battery. Joint VO+QK reversion is distinct from MLP reversion; the subsequently measured MLP-revert cost of 1.4 accuracy points appears in Table~\ref{tab:necessity}.}
\label{fig:necessity_appendix_ref}
\end{figure}
\noindent\textbf{Original direction battery.}
On the fully tuned chart model, removing 56 readout-selected directions across 41 heads costs 0.029 accuracy, compared with approximately zero for the three matched-random controls. The targeted-versus-untargeted concept selectivity test passes for zero of six eligible concepts. Full VO reversion costs 0.000 on the primary seed and $-0.003$ and $-0.001$ on two additional tuning seeds. Joint VO+QK reversion costs 0.009. These measurements are distinct from removal inside the VO-retained model. The original full-model direction surgeries on Flowers and EuroSAT cost 0.013 and 0.136; the EuroSAT result had no matched-random control and is not used to establish direction specificity.

\noindent\textbf{Delta structure and concept coverage.}
On Charts, mean relative Frobenius changes of the component weights are 0.006--0.009, with MLP-to-VO relative-write ratio 1.03. Effective VO relative change is 0.0105. The 56 hit directions cover eight trained concepts; leading best matches are double top (19), volatility spike (13), and rising volume (11). Per-hit alignments are 0.148--0.192 against a null percentile near 0.144. The additional ViT-B/32 chart run reaches 0.986 accuracy, has six coherent heads of 72 and a zero-head null tune, but fails the 0.20 coherent-fraction replication criterion. Its embedding-width-512 null threshold is approximately $1.22\times$ the width-768 threshold in these runs.

\noindent\textbf{Cross-corpus readout.}
The four LAION-2B hits have alignments 0.158--0.231 against a null threshold of 0.151, with labels green candlestick, bearish candlestick, head-and-shoulders, and ascending triangle. Missing per-step competitor traces limit their semantic disambiguation. The reported count is the number of hits passing the readout criteria.

\subsection{Continuous metrics and recovery intervals}
\label{app:continuous}
\begin{table}[htbp]
\centering
\caption{Complete continuous-metric results with paired-bootstrap 95\% intervals. $N_L$ and $N_M$ are raw loss and margin costs; $R_L$ and $R_M$ are normalized recovery. Every condition uses the same fixed logit scale within a checkpoint.}
\label{tab:continuous}
\small
\setlength{\tabcolsep}{4pt}
\begin{tabular}{@{}lccccc@{}}
\toprule
Setting & Group & $N_L$ & $N_M$ & $R_L$ & $R_M$ \\
\midrule
Charts & VO & -0.008 [-0.025, 0.006] & 1.519 [1.450, 1.582] & 0.746 [0.736, 0.757] & 0.283 [0.275, 0.291] \\
 & QK & 0.024 [0.013, 0.037] & 0.946 [0.901, 0.992] & 0.466 [0.453, 0.480] & 0.147 [0.143, 0.152] \\
 & MLP & 0.187 [0.160, 0.214] & 5.082 [4.941, 5.221] & 0.988 [0.982, 0.994] & 0.687 [0.680, 0.693] \\
Flowers102 & VO & 0.040 [0.025, 0.053] & 1.458 [1.399, 1.516] & 0.631 [0.611, 0.655] & 0.259 [0.256, 0.263] \\
 & QK & 0.009 [0.004, 0.014] & 1.022 [0.989, 1.053] & 0.241 [0.222, 0.263] & 0.157 [0.151, 0.162] \\
 & MLP & 0.160 [0.130, 0.188] & 3.226 [3.139, 3.316] & 0.880 [0.860, 0.902] & 0.545 [0.540, 0.550] \\
EuroSAT & VO & 0.004 [-0.015, 0.021] & 2.940 [2.844, 3.029] & 0.822 [0.803, 0.843] & 0.225 [0.221, 0.229] \\
 & QK & -0.008 [-0.020, 0.005] & 2.138 [2.076, 2.201] & 0.464 [0.447, 0.483] & 0.136 [0.132, 0.141] \\
 & MLP & 0.055 [0.026, 0.081] & 7.466 [7.315, 7.613] & 0.980 [0.964, 0.999] & 0.570 [0.564, 0.576] \\
Aircraft & VO & 0.202 [0.171, 0.234] & 0.981 [0.907, 1.058] & 0.505 [0.488, 0.523] & 0.267 [0.257, 0.278] \\
 & QK & 0.099 [0.076, 0.122] & 0.612 [0.559, 0.665] & 0.233 [0.216, 0.250] & 0.102 [0.092, 0.112] \\
 & MLP & 0.366 [0.325, 0.407] & 1.721 [1.615, 1.821] & 0.698 [0.675, 0.722] & 0.470 [0.454, 0.486] \\
Flowers102 (B/16) & VO & 0.115 [0.095, 0.136] & 1.363 [1.286, 1.435] & 0.486 [0.470, 0.502] & 0.239 [0.235, 0.243] \\
 & QK & 0.019 [0.014, 0.024] & 0.571 [0.542, 0.601] & 0.113 [0.103, 0.124] & 0.087 [0.082, 0.091] \\
 & MLP & 0.464 [0.417, 0.517] & 3.675 [3.561, 3.791] & 0.846 [0.828, 0.862] & 0.632 [0.623, 0.640] \\
\bottomrule
\end{tabular}
\end{table}

\begin{table}[htbp]
\centering
\caption{Accuracy recovery with paired-bootstrap 95\% intervals, including QK-only retention.}
\label{tab:recoveryci}
\small
\setlength{\tabcolsep}{4pt}
\begin{tabular}{@{}lccc@{}}
\toprule
Setting & VO & QK & MLP \\
\midrule
Charts & 0.808 [0.775, 0.839] & 0.337 [0.302, 0.373] & 0.989 [0.979, 0.998] \\
Flowers102 & 0.547 [0.497, 0.598] & 0.202 [0.159, 0.246] & 0.862 [0.824, 0.899] \\
EuroSAT & 0.869 [0.846, 0.892] & 0.354 [0.323, 0.384] & 0.960 [0.946, 0.973] \\
Aircraft & 0.428 [0.390, 0.465] & 0.187 [0.157, 0.219] & 0.682 [0.643, 0.720] \\
Flowers102 (B/16) & 0.438 [0.397, 0.481] & 0.132 [0.101, 0.165] & 0.808 [0.773, 0.843] \\
\bottomrule
\end{tabular}
\end{table}

The cross-entropy and margin tables use the model's unchanged logit scale for every condition within a checkpoint. Loss and margin costs are in their raw units; normalized recovery divides by the corresponding base-to-full improvement. Differences in raw margins between model architectures should therefore be interpreted within their own fixed-scale evaluations. Bootstrap intervals quantify evaluation-example variation conditional on each trained checkpoint through joint resampling.

\begin{table}[htbp]
\centering
\caption{Agreement on the fixed pretrained-error subset. Both-correct counts joint corrections; same-wrong counts shared incorrect predictions. Agreement includes both categories.}
\label{tab:erroragreement}
\small
\setlength{\tabcolsep}{4pt}
\begin{tabular}{@{}lccccc@{}}
\toprule
Setting & Pair & $n$ & Agreement [95\% CI] & Both correct & Same wrong \\
\midrule
Charts & VO/full & 674 & 0.7982 [0.7685, 0.8279] & 532 & 6 \\
 & MLP/full & 674 & 0.9837 [0.9733, 0.9926] & 653 & 10 \\
 & VO/MLP & 674 & 0.8056 [0.7745, 0.8353] & 532 & 11 \\
Flowers102 & VO/full & 426 & 0.5493 [0.5000, 0.5962] & 223 & 11 \\
 & MLP/full & 426 & 0.8521 [0.8192, 0.8850] & 352 & 11 \\
 & VO/MLP & 426 & 0.6714 [0.6244, 0.7160] & 222 & 64 \\
EuroSAT & VO/full & 1061 & 0.8577 [0.8360, 0.8784] & 882 & 28 \\
 & MLP/full & 1061 & 0.9491 [0.9359, 0.9623] & 973 & 34 \\
 & VO/MLP & 1061 & 0.8662 [0.8454, 0.8860] & 864 & 55 \\
Aircraft & VO/full & 1379 & 0.3916 [0.3677, 0.4170] & 361 & 179 \\
 & MLP/full & 1379 & 0.6186 [0.5917, 0.6439] & 608 & 245 \\
 & VO/MLP & 1379 & 0.5751 [0.5475, 0.6012] & 345 & 448 \\
Flowers102 (B/16) & VO/full & 606 & 0.4604 [0.4208, 0.5000] & 243 & 36 \\
 & MLP/full & 606 & 0.8053 [0.7723, 0.8366] & 450 & 38 \\
 & VO/MLP & 606 & 0.5875 [0.5479, 0.6254] & 241 & 115 \\
\bottomrule
\end{tabular}
\end{table}

The agreement analysis conditions on the fixed pretrained-error subset before resampling. The joint-correction fractions equal the VO/MLP ``both correct'' counts divided by that subset's size. For Aircraft, for example, 345 of 1,379 pretrained errors are corrected by both retentions, while 448 yield the same incorrect prediction. This separates useful overlap from shared errors.

\section{Analysis Protocol and Decision Record}
\label{app:protocol}
This record distinguishes original decision rules from analyses introduced during revision. It summarizes the available protocol and outcome information; the accompanying analysis configuration records prior exposure to the earlier accuracy results. Internal specification is not third-party-timestamped pre-registration.

\noindent\textbf{Original chart audit.}
A valid tune had to reach at least 0.60 accuracy. The original direction-necessity gate required a drop of at least 15 accuracy points, at least twice the matched-random drop, and at least twofold selectivity on targeted concepts. The observed 2.9-point drop and failed selectivity do not meet that compound gate. Semantic attribution used a threefold real/null contrast, mean cross-head absolute cosine at most 0.80, and coverage of at least five concepts. Counts of 41 versus zero, cosine 0.175, and coverage of eight satisfy the substantive attribution checks; a multiplicative ratio to zero is not reported as a finite effect size.

\noindent\textbf{Original standard-data dissociation gate.}
Each valid tune required an accuracy gain of at least 0.20. The joint dissociation prediction required VO-revert cost at most 0.05, VO-only recovery at most 0.25, and MLP-only recovery at least 0.50, with the pattern required in at least two standard datasets. All VO recoveries exceed 0.25, so the dissociation hypothesis fails. The selection of Aircraft in place of CUB-200 was attributed to availability before the corresponding runs; EuroSAT replaced Pets because the latter's reported zero-shot accuracy left insufficient headroom for the gain criterion.

\noindent\textbf{Revision-stage component and direction analysis.}
The missing necessity cells, independent LayerNorm retention conditions, and paired intervals complete the component evidence. The five-point numerical bound is extended to the component matrix during revision; it was not an original all-component threshold. The direction-removal gate required at least 50\% of VO-retained recovery and twice the mean random collapse, using the implemented denominator safeguard for near-zero random collapse. Only EuroSAT passes. The confidence intervals on identified-minus-random collapse are conditional on three controls, with seed ranges reported in Table~\ref{tab:direction}.

\noindent\textbf{Loss, margin, and random retention.}
The new analysis configuration states that earlier accuracy results had already been seen. It specifies cross-entropy and margin evaluation, ten count-matched random masks, and direct VO weights-only retention. Its results reproduce earlier VO and MLP accuracy recoveries to four decimals. Agreement is computed by post-processing the saved predictions. These are follow-up analyses of the existing checkpoints, designed with the earlier accuracy results available.

\section{Reproducing the Numerical Tables}
\label{app:reproducibility}
From the root of the accompanying source and analysis package, install NumPy and run:
\begin{verbatim}
python -m pip install numpy
python scripts/build_tables.py
latexmk -pdf -interaction=nonstopmode -halt-on-error main.tex
\end{verbatim}
The final command requires a LaTeX installation with \texttt{latexmk} and BibTeX. The analysis runs on CPU from saved outputs; it requires neither a GPU nor model downloads.

The five \texttt{data/persample\_*.npz} files contain labels and condition-level predictions. The script recomputes accuracy necessity, normalized recovery, paired-bootstrap intervals (2,000 joint resamples, seed 0), and per-class evaluation counts from these arrays. Continuous-metric tables use the recorded estimates and intervals in \texttt{data/SIXRSF\_results.json}; agreement tables use \texttt{data/full\_model\_agreement.csv}. These tables therefore expose saved-result summaries as well as independently recomputed accuracy contrasts. The configuration in \texttt{data/} documents the follow-up analysis.

This source-package workflow regenerates eight component-analysis tables and the associated accuracy and class-count JSON files. The separate reproduction package also includes the historical semantic-readout and direction-intervention JSON results, plus the original training, data-generation, and intervention scripts. From its root, the CPU verification commands are:
\begin{verbatim}
python -m pip install numpy
python scripts/rebuild_tables.py
python runs/analyze_agreement.py
\end{verbatim}
The first analysis command rebuilds four result summaries from saved JSON files and checks consistency across result files. The second recomputes prediction agreement and joint corrections from per-example predictions. These CPU workflows have been checked on the distributed files. GPU training and intervention runs were not repeated during packaging.

For model-level replication, the reproduction README provides the environment and script sequence. Charts are generated locally; standard datasets are obtained through the included loaders. The archive supplies checkpoint hashes but not the checkpoint weights. The chart and standard-data training scripts produce the required tuned models; intervention scripts subsequently load them. Repeated training can vary with the compute environment. The original protocol's decision history is summarized in Appendix~\ref{app:protocol}; the supplementary archive includes the sanitized full protocol log, master plan, and a record of redactions and retrospective clarifications in its \texttt{protocol/} directory.

\end{document}